# A Complete Analysis of the $\ell_{1,p}$ Group-Lasso

**Julia E. Vogt** JULIA.VOGT@UNIBAS.CH
Computer Science Department, University of Basel, Switzerland

**Volker Roth** VOLKER.ROTH@UNIBAS.CH
Computer Science Department, University of Basel, Switzerland

## Abstract

The Group-Lasso is a well-known tool for joint regularization in machine learning methods. While the $\ell_{1,2}$ and the $\ell_{1,\infty}$ version have been studied in detail and efficient algorithms exist, there are still open questions regarding other $\ell_{1,p}$ variants. We characterize conditions for solutions of the $\ell_{1,p}$ Group-Lasso for *all* $p$-norms with $1 \leq p \leq \infty$, and we present a unified active set algorithm. For all $p$-norms, a highly efficient projected gradient algorithm is presented. This new algorithm enables us to compare the prediction performance of many variants of the Group-Lasso in a *multi-task* learning setting, where the aim is to solve many learning problems in parallel which are coupled via the Group-Lasso constraint. We conduct large-scale experiments on synthetic data and on two real-world data sets. In accordance with theoretical characterizations of the different norms we observe that the weak-coupling norms with $p$ between 1.5 and 2 consistently outperform the strong-coupling norms with $p \gg 2$.

## 1. Introduction

In 1996, (Tibshirani, 1996) introduced the Lasso, an $\ell_1$-constrained method for sparse variable selection. This method was extended by (Yuan & Lin, 2006) and by (Turlach et al., 2005) to the problem where explanatory factors are represented as *groups* of variables, leading to solutions that are sparse on the group level. In recent years, mainly two variants of the Group-Lasso have been proposed: one uses the $\ell_{1,2}$ norm and the other one the $\ell_{1,\infty}$ norm for regularization. Many algorithms for the $\ell_{1,2}$-version have been presented, see for instance (Yuan & Lin, 2006; Meier et al., 2008; Argyriou et al., 2007; Kim et al., 2006) or (Bach, 2008). Algorithms for the $\ell_{1,\infty}$-variant of the Group-Lasso were studied in (Turlach et al., 2005; Schmidt & Murphy, 2008; Quattoni et al., 2009) and (Vogt & Roth, 2010). The mixed-norm regularization was elaborated in (Liu & Ye, 2010) and (Zhang et al., 2010). In (Liu & Ye, 2010), an $\ell_{1,p}$-regularized Euclidean projection is presented and the optimization problem is solved through an accelerated gradient method. However, for large-scale problems with thousands of groups, this method is not efficient.

In this work we derive conditions for the *completeness* and *uniqueness* of all $\ell_{1,p}$ Group-Lasso estimates, where a solution is *complete*, if it includes all groups that might be relevant in other solutions. Based on these conditions it can easily be tested if a solution is complete, and all other groups that may be included in alternative solutions with identical costs can be identified. We show the efficiency of this active set algorithm and we prove convergence to the global optimizer.

Our main technical contribution in this work is threefold: i) We present a unified characterization of solutions for *all* $\ell_{1,p}$ Group-Lasso methods. ii) We present an efficient nesting of a constrained optimization problem and a Lagrangian optimization problem. During optimization, we use a projected gradient method that works with the Lagrangian form of an optimization problem. However, the active set algorithm needs the constrained form of the problem. The efficient combination of these two optimization problems is not trivial, as finding the Lagrangian parameter can be arbitrarily sensitive to the step length which implies slow convergence of the algorithm. We show that we can combine these two methods efficiently by using an interval bisection for finding the Lagrangian parameter that is guaranteed to converge. iii) These new theoretic and algorithmic developments allow us to conduct large-scale comparison experiments between various different $\ell_{1,p}$ Group-Lasso versions. We fo-cus on *multi-task* (or *transfer*) learning problems, in which the individual tasks are coupled via the group-structure of the constraint term. The underlying assumption here is that multiple tasks share a common sparsity pattern. Large-scale experiments reveal clear and statistical significant differences in the prediction performance of the different $\ell_{1,p}$ methods. Theoretical analysis shows a direct relation between norms with high $p$-values and increasing coupling strength of the Group-Lasso constraint.

## 2. Characterization of Solutions for the $\ell_{1,p}$ Group-Lasso

We consider the following setting of a generalized linear model (see (McCullagh & Nelder, 1983) for more details): given an i.i.d. data sample $\{\boldsymbol{x}_1, ..., \boldsymbol{x}_n\}$, $\boldsymbol{x}_i \in \mathbb{R}^d$, arranged as rows of the data matrix $X$, and a corresponding vector of responses $\boldsymbol{y} = (y_1, ..., y_n)^T$, we want to minimize the negative log-likelihood

$$l(\boldsymbol{y}, \boldsymbol{\nu}, \theta) = -\sum_i \log f(y_i; \nu_i, \theta), \tag{1}$$

where the exponential-familiy distribution $f$ is the random component of a generalized linear model (GLM),

$$f(y; \nu, \theta) = \exp(\theta^{-1}(y\nu - b(\nu)) + c(y, \theta)). \tag{2}$$

The GLM is completed by introducing a systematic component $\nu = \boldsymbol{x}^T\boldsymbol{\beta}$ and a strictly monotone differentiable (canonical) link function specifying the relationship between the random and systematic components: $\eta(\mu) = \nu$, where $\mu = E_\nu[y]$ is related to the natural parameter $\nu$ of the distribution $f$ by $\mu = b'(\nu) = \eta^{-1}(\nu)$. From a technical perspective, an important property of this framework is that $\log f(y; \nu, \theta)$ is strictly concave in $\nu$. This follows from the fact that the sufficient statistics $y/\theta$ is one-dimensional and, therefore, *minimal*, which implies that the log partition function $b(\nu)/\theta$ is strictly convex, see (Brown, 1986). For the sake of simplicity we fix the scale parameter $\theta$ to 1. With $\nu = \boldsymbol{x}^T\boldsymbol{\beta}$ and $b'(\nu) = \eta^{-1}(\nu)$, the gradient can be seen as a function in either $\boldsymbol{\nu}$ or $\boldsymbol{\beta}$:

$$\nabla_{\boldsymbol{\nu}} l(\boldsymbol{\nu}) = -(\boldsymbol{y} - \eta^{-1}(\boldsymbol{\nu})) \text{ or} \tag{3}$$

$$\nabla_{\boldsymbol{\beta}} l(\boldsymbol{\beta}) = -X^T\nabla_{\boldsymbol{\nu}} l(\boldsymbol{\nu}) = -X^T(\boldsymbol{y} - \eta^{-1}(X^T\boldsymbol{\beta})), \tag{4}$$

For the following analysis, we partition $X$, $\boldsymbol{\beta}$ and $\boldsymbol{h} := \nabla_{\boldsymbol{\beta}} l$ into $J$ subgroups: $X = (X_1, ..., X_J)$,

$$\boldsymbol{\beta} = \begin{pmatrix} \boldsymbol{\beta}_1 \\ \vdots \\ \boldsymbol{\beta}_J \end{pmatrix}, \quad \boldsymbol{h} = \begin{pmatrix} \boldsymbol{h}_1 \\ \vdots \\ \boldsymbol{h}_J \end{pmatrix} = \begin{pmatrix} X_1^\top \nabla_{\boldsymbol{\nu}} l \\ \vdots \\ X_J^\top \nabla_{\boldsymbol{\nu}} l \end{pmatrix}. \tag{5}$$

$l$ is a strictly convex function in $\boldsymbol{\nu}$. For general matrices $X$ it is convex in $\boldsymbol{\beta}$, and it is strictly convex in $\boldsymbol{\beta}$ if $X$ has full rank and $d \le n$. Given $X$ and $\boldsymbol{y}$, the Group-Lasso minimizes the negative log-likelihood viewed as a function in $\boldsymbol{\beta}$ under a constraint on the sum of the $\ell_p$-norms, $1 \le p \le \infty$, of the subvectors $\boldsymbol{\beta}_j$:

$$\text{minimize } l(\boldsymbol{\beta}) \quad \text{s.t.} \quad g(\boldsymbol{\beta}) \ge 0, \tag{6}$$

$$\text{where} \quad g(\boldsymbol{\beta}) = \kappa - \textstyle\sum_{i=1}^J \|\boldsymbol{\beta}_j\|_p. \tag{7}$$

Here $g(\boldsymbol{\beta})$ is implicitly a function of the fixed parameter $\kappa$. Considering the *unconstrained* problem, the solution is not unique if the dimensionality exceeds $n$: every $\boldsymbol{\beta}^* = \boldsymbol{\beta}^0 + \boldsymbol{\xi}$ with $\boldsymbol{\xi}$ being an element of the null space $\boldsymbol{N}(X)$ is also a solution. By defining the unique value $\kappa_0 := \min_{\boldsymbol{\xi} \in \boldsymbol{N}(X)} \sum_{i=1}^J \|\boldsymbol{\beta}_j^0 + \boldsymbol{\xi}_j\|_p$, we will require that the constraint is active i.e. $\kappa < \kappa_0$. Although it might be infeasible to ensure this activeness by computing $\kappa_0$ and selecting $\kappa$ accordingly, practical algorithms will not suffer from this problem: given a solution, we can always check if the constraint was active. If this was not the case, then the uniqueness question reduces to checking if $d \le n$. In this case the solutions are usually not sparse, because the feature selection mechanism has been switched off. To produce a sparse solution, one can then try smaller $\kappa$-values until the constraint is active. We will restrict our further analysis to models with finite likelihood ($f < +\infty$), i.e. $l > -\infty$. Technically this means that we require that the domain of $l$ is $\mathbb{R}^d$, which implies that Slater's condition holds.

**Theorem 2.1** *If $\kappa < \kappa_0$ and $X$ has maximum rank, then the following holds: (i) A solution $\widehat{\boldsymbol{\beta}}$ exists and $\sum_{i=1}^J \|\widehat{\boldsymbol{\beta}}_j\|_p = \kappa$ for any such solution. (ii) If $d \le n$, the solution is unique.*

*Proof:* Under the assumption $l > -\infty$ a minimum of (6) is guaranteed to exist, since $l$ is continuous and the region of feasible vectors $\beta$ is compact. Since we assume that the constraint is active, any solution $\widehat{\boldsymbol{\beta}}$ will lie on the boundary of the constraint region. It is easily seen that $\sum_{j=1}^J \|\boldsymbol{\beta}_j\|_p$ is convex for $1 \le p \le \infty$ which implies that $g(\boldsymbol{\beta})$ is concave. Thus, the region of feasible values defined by $g(\boldsymbol{\beta}) \ge 0$ is convex. If $d \le n$, the objective function $l$ will be strictly convex, which implies that the minimum is unique. □

The Lagrangian for problem (6) reads

$$\mathcal{L}(\boldsymbol{\beta}, \lambda) = l(\boldsymbol{\beta}) - \lambda g(\boldsymbol{\beta}). \tag{8}$$

For a given $\lambda > 0$, $\mathcal{L}(\boldsymbol{\beta}, \lambda)$ is a convex function in $\boldsymbol{\beta}$. The vector $\widehat{\boldsymbol{\beta}}$ minimizes $\mathcal{L}(\boldsymbol{\beta}, \lambda)$ iff the $d$-dimensional null-vector $\boldsymbol{0}_d$ is an element of the subdifferential $\partial_{\boldsymbol{\beta}}\mathcal{L}(\boldsymbol{\beta}, \lambda)$. The subdifferential is

$$\partial_{\boldsymbol{\beta}}\mathcal{L}(\boldsymbol{\beta}, \lambda) = \nabla_{\boldsymbol{\beta}} l(\boldsymbol{\beta}) + \lambda \boldsymbol{v} = X^\top \nabla_{\boldsymbol{\nu}} l(\boldsymbol{\nu}) + \lambda \boldsymbol{v}, \tag{9}$$

with $\boldsymbol{v} = (\boldsymbol{v}_1, \dots \boldsymbol{v}_J)^\top$ defined by

$$\begin{aligned} \|\boldsymbol{v}_j\|_q \leq 1 \quad \text{if} \quad \|\boldsymbol{\beta}_j\|_p = 0 \\ \|\boldsymbol{v}_j\|_q = 1 \quad \text{if} \quad \|\boldsymbol{\beta}_j\|_p > 0, \end{aligned} \tag{10}$$

where $\frac{1}{p} + \frac{1}{q} = 1$ for $1 < p < \infty$ and if $p = 1$, then $q = \infty$ and vice versa. Thus, $\widehat{\boldsymbol{\beta}}$ is a minimizer for fixed $\lambda$ iff $\mathbf{0}_d = -X^T \nabla_{\boldsymbol{\nu}} l(\boldsymbol{\nu})|_{\boldsymbol{\nu}=\widehat{\boldsymbol{\nu}}} + \lambda \boldsymbol{v}$, (with $\widehat{\boldsymbol{\nu}} = X\widehat{\boldsymbol{\beta}}$). Let $d_j$ denote the dimension of the $j$-th subvector $\boldsymbol{\beta}_j$. Hence, for all $j$ with $\widehat{\boldsymbol{\beta}_j} = \mathbf{0}_{d_j}$ it holds that $\lambda \geq \|X_j^T \nabla_{\boldsymbol{\nu}} l(\boldsymbol{\nu})|_{\boldsymbol{\nu}=\widehat{\boldsymbol{\nu}}}\|_q$. This yields $\lambda = \max_j \|X_j^T \nabla_{\boldsymbol{\nu}} l(\boldsymbol{\nu})|_{\boldsymbol{\nu}=\widehat{\boldsymbol{\nu}}}\|_q$ and for all $j$ with $\widehat{\boldsymbol{\beta}_j} \neq \mathbf{0}_{d_j}$ it holds that $\lambda = \|X_j^T \nabla_{\boldsymbol{\nu}} l(\boldsymbol{\nu})|_{\boldsymbol{\nu}=\widehat{\boldsymbol{\nu}}}\|_q$ .

With these derivations we obtain results analogous to (Roth & Fischer), but for *all* $p$-norms: Assume that an algorithm has found a solution $\widehat{\boldsymbol{\beta}}$ of (6) with the set of "active" groups $\mathcal{A} := \{j : \widehat{\boldsymbol{\beta}}_j \neq \mathbf{0}\}$. If $\mathcal{A} = \mathcal{B} = \{j : \|\widehat{\boldsymbol{h}}_j\|_q = \lambda\}$, then there cannot exist any other solution with an active set $\mathcal{A}'$ with $|\mathcal{A}'| > |\mathcal{A}|$. Thus, $\mathcal{A} = \mathcal{B}$ implies that the solution is complete. Otherwise, the additional elements in $\mathcal{B}$ which are not contained in $\mathcal{A}$ define all possible groups that potentially become active in alternative solutions. However, $\mathcal{A}$ might still contain redundant groups. There exists a simple test procedure for uniqueness under a further rank assumption of the data matrix $X$ (Roth & Fischer): If every $n \times n$ submatrix of $X$ has full rank, $\mathcal{A}$ denotes the active set corresponding to some solution $\widehat{\boldsymbol{\beta}}$ of (6) and $X_{\mathcal{A}}$ denotes the $n \times s$ submatrix of $X$ composed of all active groups. Then, if $\mathcal{A}$ is complete and if $s \leq n$, $\widehat{\boldsymbol{\beta}}$ is the unique solution of (6). Figure 1 shows a graphical representation of different $\ell_p$ norms.

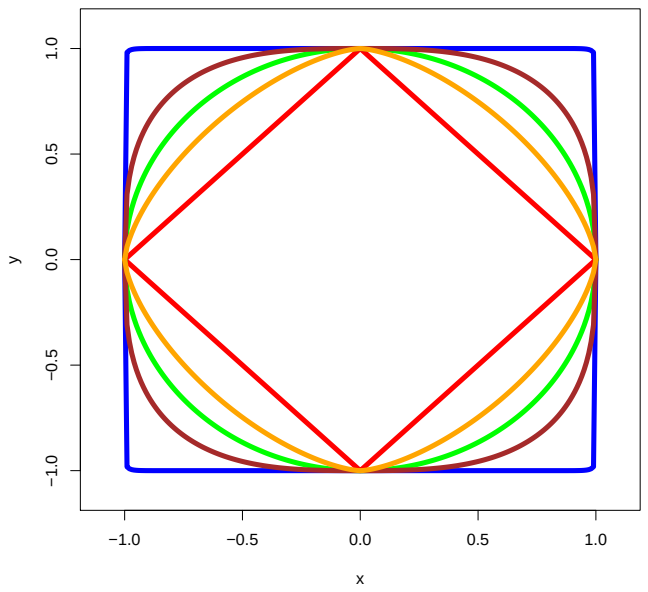

*Figure 1.* Different $\ell_p$ balls: red curve: $\ell_1$, orange curve: $\ell_{1.5}$, green curve: $\ell_2$, brown curve: $\ell_3$, blue curve: $\ell_\infty$.

## 3. An Efficient Active-Set Algorithm

The characterization of the optimal solution presented in section 2 allows us to construct an active set algorithm to solve the constrained optimization problem (6) for all $\ell_{1,p}$ norms. The algorithm starts with only one active group. In every iteration, further active groups are selected or removed, depending on the violation of the Lagrangian condition. The algorithm is a straightforward generalization of the subset algorithm for the standard Lasso problem presented in (Osborne et al., 2000). The main idea is to find a small set of active groups. The optimization in step **B** can be performed by the projected gradient method (Bertsekas, 1995). The main challenge typically is to compute efficient projections onto the $\ell_{1,p}$ ball. In general this is a hard to solve nonlinear optimization problem with nonlinear and even non-differentiable constraints. For the $\ell_{1,2}$ norm, (Kim et al., 2006) presented an efficient algorithm for the projection to the $\ell_{1,2}$ ball and the projection to the $\ell_{1,\infty}$ ball can be performed efficiently by the method introduced in (Quattoni et al., 2009). The $\ell_{1,1}$ ball can be seen as a special case of the projection to the $\ell_{1,2}$ ball. An efficient projection to the $\ell_{1,p}$ ball was presented in (Liu & Ye, 2010). In general, the main idea in the projected gradient method is that one does not optimize problem (6) directly but solves a subproblem with quadratic cost instead. First, we take a step $s\nabla_{\boldsymbol{\beta}} l(\boldsymbol{\beta})$ along the the negative gradient with step size $s$ and obtain the vector $\boldsymbol{b} = \boldsymbol{\beta} - s\nabla_{\boldsymbol{\beta}} l(\boldsymbol{\beta})$. We then project $\boldsymbol{b}$ on the convex feasible region to obtain a feasible vector. Hence, the minimization problem we need to solve now reads

**Algorithm 1** Active Set Algorithm

**A** : Set $\mathcal{A} = j_0$, $\boldsymbol{\beta}_{j_0}$ arbitrary with $\|\boldsymbol{\beta}_{j_0}\|_p = \kappa$.

**B** : **Optimize** over the current active set $\mathcal{A}$.

Define set $\mathcal{A}^+ = \left\{ j \in \mathcal{A} : \|\boldsymbol{\beta}_{j_0}\|_p > 0 \right\}$.
Define $\lambda = \max_{j \in \mathcal{A}^+} \|\boldsymbol{h}_j\|_q$. Adjust the active set $\mathcal{A} = \mathcal{A}^+$.

**C** : **Lagrangian violation**: $\forall j \notin \mathcal{A}$, check if $\|\boldsymbol{h}_j\|_q \leq \lambda$. If this is the case, we have found a global solution. Otherwise, include the group with the largest violation to $\mathcal{A}$ and go to **B**.

**D: Check for completeness and uniqueness.**

$$\min_{\boldsymbol{\beta}} \quad \|\boldsymbol{b} - \boldsymbol{\beta}\|_2^2 + \mu \Big( \sum_{j=1}^{J} \|\boldsymbol{\beta}_j\|_p - \kappa \Big) \tag{11}$$

with Lagrangian multiplier $\mu$. Algorithm 2 shows the projection for all $\ell_{1,p}$ norms with $1 < p < \infty$.

**Convergence of Interval Bisection.** It remains to show that the interval bisection within Algorithm 2 converges. This is our main technical contribution in this work: the efficient combination of a constrained problem with the Lagrangian form of an optimization problem. The projection algorithm proposed in (Liu & Ye, 2010) needs the Lagrangian representation of the problem while we work with the constrained form in the active set algorithm. The combination of these two

**Algorithm 2** Optimization Step B

**B1 : Gradient :**
At time $t-1$, set $\boldsymbol{b} = \boldsymbol{\beta}^{t-1} - s\nabla_{\boldsymbol{\beta}} l(\boldsymbol{\beta}^{t-1})$ and $\mathcal{A}^+ = \mathcal{A}$, where $s$ is the step size parameter.
**Initialize** Lagrangian multiplier $\mu$ within the interval $(0, \mu_{\max})$.

**B2 : Projection :**
For all $j \in \mathcal{A}^+$ minimize (11):
**While** $\sum_{j=1}^{J} \|\boldsymbol{\beta}_j^t\|_p \neq \kappa$ **do**
*Compute* projection as in (Liu & Ye, 2010), obtain optimal $\boldsymbol{\beta}_j^*$ for all $j \in \mathcal{A}^+$. *Adapt* Lagrangian multiplier $\mu$ via interval bisection.
**B3 : New solution:** $\forall j \in \mathcal{A}^+$, set $\boldsymbol{\beta}_j^t = \boldsymbol{\beta}_j^*$

optimization problems is not trivial, as finding the appropriate Lagrangian multiplier $\mu$ could be arbitrarily sensitive to the step length $s$ what leads to extremely slow convergence of the algorithm. Our contribution is to show that we can combine these two methods by using an interval bisection for finding the Lagrangian parameter $\mu$ that is guaranteed to converge rapidly.

**Theorem 3.1** *The interval bisection in Algorithm 2 is guaranteed to converge.*

To prove Theorem 3.1, we need the following Lemma:

**Lemma 3.2** *For two Lagrangian functions with convex likelihood function $f(\boldsymbol{\beta})$*

$$\mathcal{L}_1(\boldsymbol{\beta}, \mu_1) := f(\boldsymbol{\beta}) + \mu_1\Big(\sum_{j=1}^{J} \left\|\boldsymbol{\beta}_j\right\|_p - \kappa_1\Big) \text{ and}$$

$$\mathcal{L}_2(\boldsymbol{\beta}, \mu_2) := f(\boldsymbol{\beta}) + \mu_2\Big(\sum_{j=1}^{J} \left\|\boldsymbol{\beta}_j\right\|_p - \kappa_2\Big)$$

*it holds that:* $\mu_1 < \mu_2 \iff \kappa_2 < \kappa_1$.

The proof of Lemma 3.2 is done via perturbation and sensitivity analysis (see e.g. (Forst & Hoffmann, 2010) or (Bertsekas, 1995) for more details) and is presented in the supplementary material. Now we can prove Theorem 3.1:

*Proof:* Let $\tilde{g}(\mu) := \sum_{j=1}^{J} \left\|\boldsymbol{\beta}_j(\mu)\right\|_p - \kappa$. We denote with $\boldsymbol{\beta}(\mu) := \arg\min_{\boldsymbol{\beta}} \mathcal{L}(\boldsymbol{\beta}, \mu)$ the optimal $\boldsymbol{\beta}$ for the Lagrangian function $\mathcal{L}(\boldsymbol{\beta}, \mu)$ as defined in Lemma 3.2. Then we get with Lemma 3.2 and because we know that the solution lies on the boundary of the feasible set for $\mu_1 < \mu_2$:

$$\tilde{g}(\mu_1) = \underbrace{\sum_{j=1}^{J} \left\|\boldsymbol{\beta}_j(\mu_1)\right\|_p}_{=\kappa_1} - \kappa > \underbrace{\sum_{j=1}^{J} \left\|\boldsymbol{\beta}_j(\mu_2)\right\|_p}_{=\kappa_2} - \kappa = \tilde{g}(\mu_2).$$

Hence $\tilde{g}$ is a monotonically decreasing and continuous function in the interval $[0, \mu_{\max}]$ where $\mu_{\max} := \|\boldsymbol{\beta}\|_q$ (see (Liu & Ye, 2010) for details about $\mu_{\max}$). For $f(\boldsymbol{\beta}) := \|\boldsymbol{b} - \boldsymbol{\beta}\|_2^2$ it holds that $\tilde{g}(0) = \sum_{j=1}^{J} \|\boldsymbol{b}_j\|_p - \kappa > 0$ (since we assume that the constraint is active) and $\tilde{g}(\mu_{\max}) = \sum_{j=1}^{J} \|\boldsymbol{0}\|_p - \kappa < 0$ (see (Liu & Ye, 2010), Theorem 1). According to the Intermediate Value Theorem, $\tilde{g}(\mu)$ has a unique root in $(0, \mu_{\max})$, hence the interval bisection converges. □

After each iteration of the bisection method, the bounds containing the root decrease by a factor of two. As the interval bisection is guaranteed to converge, we know that we will achieve a given tolerance in the solution in a logarithmic number of iterations (see e.g. (Press et al., 2007) for more details). The convergence of the active set algorithm follows immediately: if the solution is not optimal, the solution of the augmented system will be a descent direction for the augmented problem and also for the whole problem, as primal feasibility is maintained and the constraint qualifications are fulfilled. This implies that the algorithm as a whole must converge. With these theoretical results we are now able to efficiently combine the active set algorithm with the projection algorithm for all $p$-norms. By using this efficient unified active set algorithm we can now look at the prediction performance of all $p$-norms for large scale experiments with thousands of features.

## 4. Multi-Task Applications

We address the problem of learning classifiers for a large number of tasks. In transfer or multi-task learning, we want to improve the generalization ability and the predictive power by solving many learning problems in parallel. Each task should benefit from the amount of data that is jointly given by all tasks and hence yield better results than examining every task individually. The motivation for using the Group-Lasso in problems of this kind is to *couple* the individual tasks via the group structure of the constraint term, based on the assumption that multiple tasks share a common sparsity pattern. Due to our efficient active set algorithm we are now able to handle data sets with thousands of features in reasonable time.

**Coupling strength of $\ell_p$ norms.** The coupling properties of the different $p$ norms have a major influence on the prediction performance of the Group-Lasso variants. The higher the value of $p$, the stronger the different tasks are coupled. For $p = 1$, the tasks within one group are barely coupled, as the $\ell_{1,1}$ regularization only induces a global coupling over all tasks. For $p = 2$ there exists an intermediate coupling of tasks within a group and for $p = \infty$ the coupling of the tasks is very strong. This is due to the fact that the $\ell_\infty$ norm only penalizes the maximum absolute entry of a group, meaning we can increase all other parameters in this group to the maximum value without changing the constraint. Hence we can assign maximum weight to every task in this group. The relation between coupling strength and value of $p$ is illustrated in Figure 2 and Figure 3.

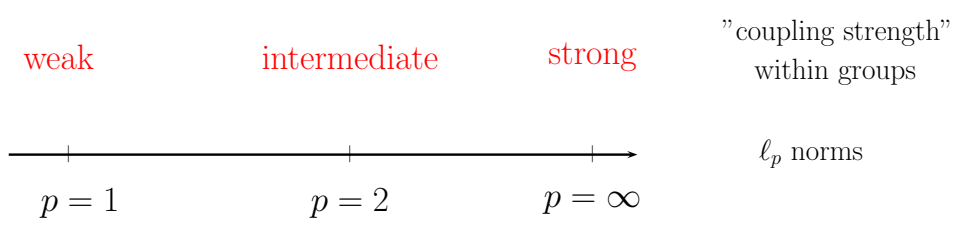

*Figure 2.* Coupling strength of $\ell_p$ norms

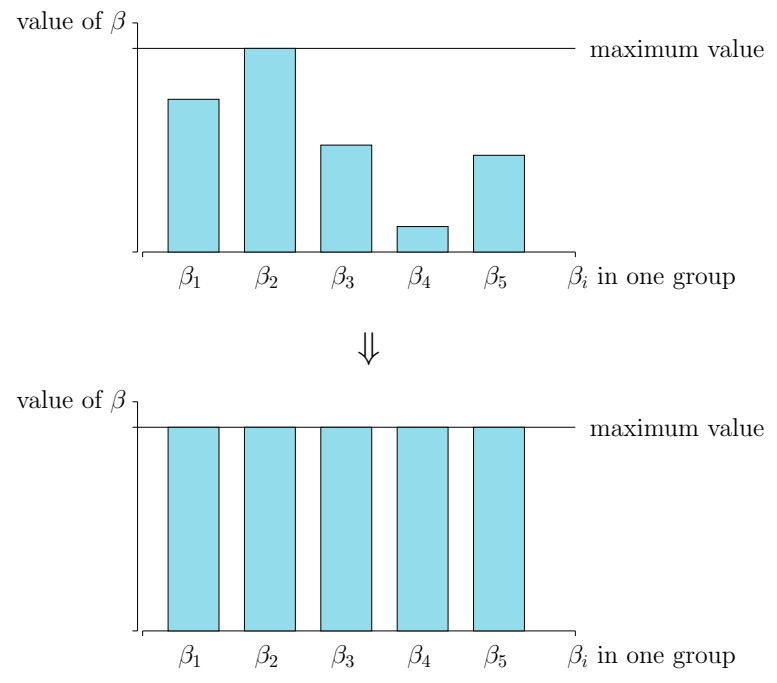

*Figure 3.* For the $\ell_{1,\infty}$ Group-Lasso, all $\beta_i$ in one group can be raised to the maximum value without changing the value of the constraint. This explains the strong coupling properties for $p = \infty$.

**Synthetic Experiments.** The synthetic data for a classification problem was created in the following way: we consider a multi-task setting with $m$ tasks and $d$ features ($\widehat{=}\ d$ groups) with a $d \times m$ parameter matrix $B = [\boldsymbol{\beta}_1, \ldots, \boldsymbol{\beta}_m]$, where $\boldsymbol{\beta}_i \in \mathbb{R}^d$ is a parameter vector for the $i$-th task. Further, assume we have a data set $D = (z_1, ..., z_n)$ with points $z$ belonging to some set $Z$, where $Z$ is the set of tuples $(\boldsymbol{x}_i, y_i, l_i)$ for $i = 1, \ldots, n$ where each $\boldsymbol{x}_i \in \mathbb{R}^d$ is a feature vector, $l_i \in 1, \ldots, m$ is a label that specifies to which of the $m$ tasks the example belongs to and $y_i \in \{-1, 1\}$ is the corresponding class label. First, we generated the parameter matrix $B$ by sampling each entry from a normal distribution $\mathcal{N}(0, 1)$. We selected 2% of the features to be the set $V$ of relevant features and zeroed the other entries.

We ran four rounds of experiments where we changed the shared sparsity pattern across the different tasks. In the first round all tasks have exactly the same sparsity pattern, just the values of $\beta_i$ differ. In the second experiment, the tasks share 75% of the sparsity pattern, in the third experiment 50% and in the last experiment only 30%. For the training set, we sampled $n$-times a $d \times m$ matrix, where each entry of the matrix was sampled from the normal distribution $\mathcal{N}(0, 1)$. The corresponding labels $\boldsymbol{y} \in \mathbb{R}^{nm}$ are computed by $\boldsymbol{y}_k = (\text{sgn}(\boldsymbol{\beta}_k^T \boldsymbol{x}_k^1), ..., \text{sgn}(\boldsymbol{\beta}_k^T \boldsymbol{x}_k^n))^T \in \mathbb{R}^n$ for $k = 1, ..., m$. The test data was obtained by splitting the training data in three parts and keeping 1/3 as an "out-of-bag" set. We fixed the number of tasks $m$ to 50, the number of features $d$ to 500 and the number of examples $n$ per task to 200.

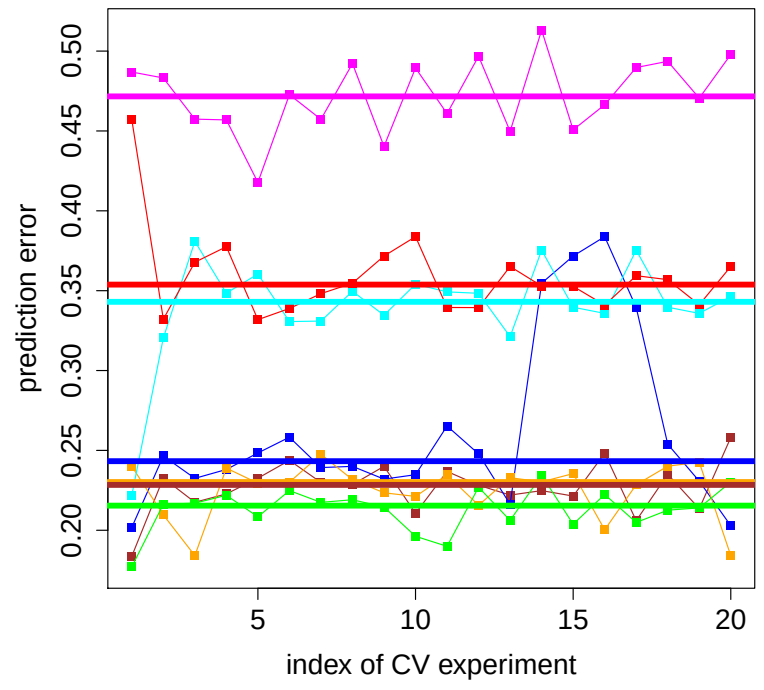

*Figure 4.* Prediction error of the different regularizers: magenta curve: learning on pooled data, red curve: single $\ell_1$, cyan curve: $\ell_{1,1}$, orange curve: $\ell_{1,1.5}$, brown curve: $\ell_{1,3}$, blue curve: $\ell_{1,\infty}$, green curve: $\ell_{1,2}$. In this Figure we have 100% shared sparsity pattern.

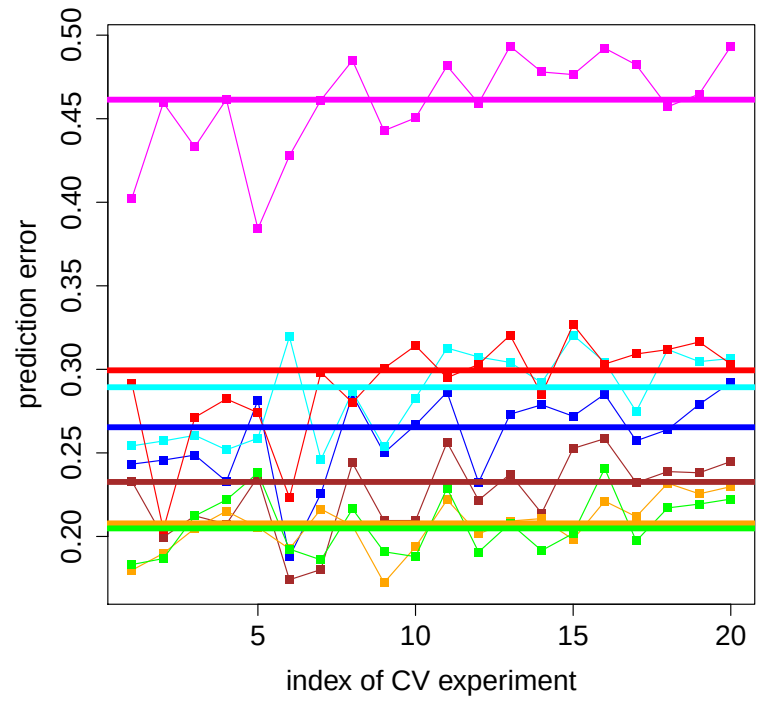

*Figure 5.* 75% shared sparsity pattern.

We compared different approaches to solve the multi-task learning problem. One approach is to pool the data, i.e. combine all tasks to one "big" task. Then we conducted single-task learning on every task separately, and we compared different $\ell_{1,p}$ Group-Lasso methods where we used the same active set algorithm, the only difference lying in the projection step. The statistical significance was tested with the Kruskal-Wallis rank-sum test for multiple testing correction and the Dunn post test with Bonferroni correction.

Figure 4 shows the result for the data set with 100% shared sparsity pattern. One can see that the pooled data performs worst and that single-task learning performs almost exactly the same as the $\ell_{1,1}$ Group-Lasso. As the $\ell_{1,1}$ norm barely couples the tasks, this result is not surprising. We perceive that single-task learning is significantly worse than multi-task learning. Between all Group-Lasso methods there is no statistical significant difference. As we have exactly the same sparsity pattern in every task, even the very strong coupling of the $\ell_{1,\infty}$ norm leads to good results.

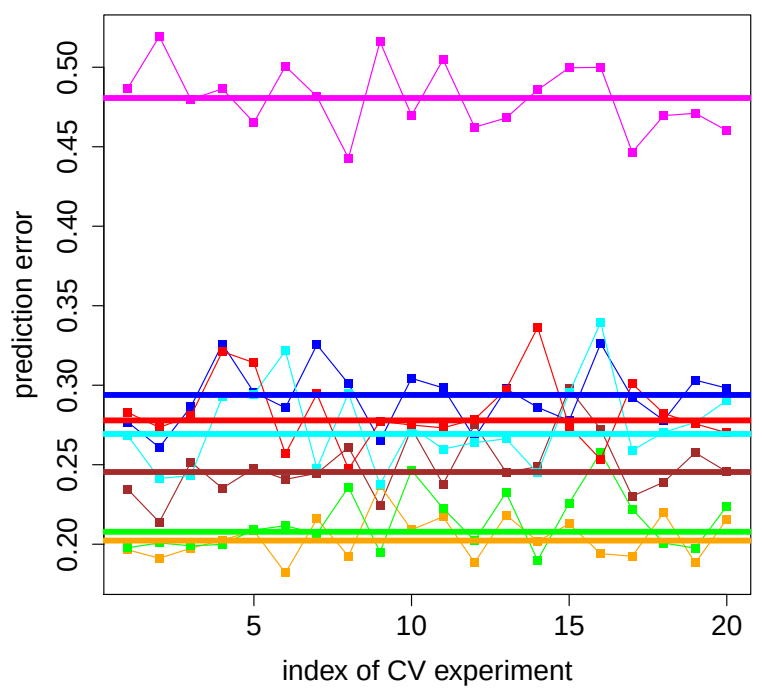

*Figure 6.* 50% shared sparsity pattern.

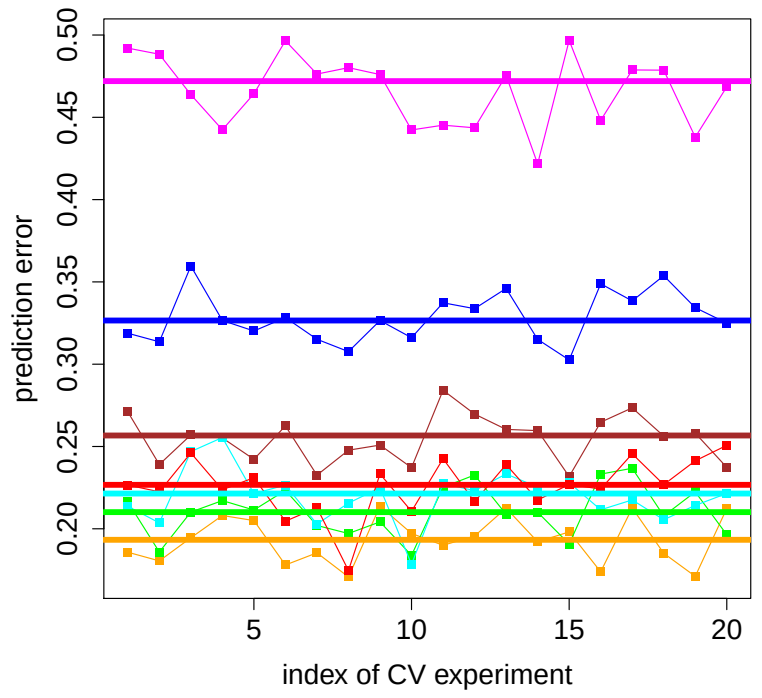

*Figure 7.* 30% shared sparsity pattern.

In Figure 5 the results for 75% shared sparsity pattern are plotted. As in the experiment with the same sparsity pattern, pooling the data is worst and multi-task learning outperforms single-task learning. Here we can see that the strong coupling of the $\ell_{1,\infty}$ norm yields worse result than in the experiment before, because the sparsity pattern is not exactly the same across the different tasks anymore. There is no significant difference between the $\ell_{1,2}$ norm and the $\ell_{1,1.5}$ norm. By further reducing the joint sparsity pattern we observe that the very tight coupling of the $\ell_{1,\infty}$ norm leads to even worse results than single-task learning and we see a statistical significant advantage of the weak coupling norms $\ell_{1,2}$ and $\ell_{1,1.5}$ over all other methods, as shown in Figure 6. If we reduce the shared sparsity pattern to only 30%, we can nicely see that in this case the weak coupling norm $\ell_{1,1.5}$ shows a clear advantage and the strong coupling norms $\ell_{1,3}$ and $\ell_{1,\infty}$ are even worse than single-task learning. These results are demonstrated in Figure 7. In all experiments, there is not one single case where the strong coupling $\ell_{1,\infty}$ norm performs better than the weak coupling regularizations. There exists a convincing explanation for the better performance of the weak coupling variants: the different tasks are connected with each other only over the constraint term. If the tasks do not share exactly the same sparsity pattern, i.e. if the model assumptions are violated, this strong coupling is sensitive to model mismatches. The $\ell_{1,\infty}$ norm couples the tasks *too* strong. For all values of $p$ with $1 \leq p \leq \infty$, values for $p \in [1.5, 2]$ seem to be the best compromise between *no* coupling and very strong coupling.

## 5. Efficiency of the Algorithm

We show the efficiency of our active set algorithm by comparing the run time of our method with the $\ell_{1,p}$ norm-regularization introduced in (Liu & Ye, 2010). To our knowledge, this is the only existing method that can compute Group-Lasso solutions for all $\ell_{1,p}$ norms. We created synthetic data in the same way as explained in section 4 and compared the run time of our algorithm and the algorithm proposed by (Liu & Ye, 2010) for a fixed number of relevant features. The code for ((Liu & Ye, 2010))'s method is publicly available[1]. The comparisons are shown in Figure 8. The dashed lines show the run time in log-log scale for the algorithm proposed in (Liu & Ye, 2010), the lines show the run time for our proposed active set algorithm. We plotted the run time for the $\ell_{1,1.5}$, $\ell_{1,3}$, $\ell_{1,\infty}$, and $\ell_{1,2}$ Group-Lasso methods in Figure 8. One can see that our active set method is by far faster if the data set contains many groups. The steep increase in the last section of (Liu & Ye, 2010)'s algorithm between 10000 and 20000 groups is due to numerical problems that arise in their optimizer by having more than 10000 groups. This comparison shows the huge advantage of using an active set method. If the solution is sparse, not all groups are selected, but only the active ones, and the active set algorithm only has to optimize over the active set, but not over the set of *all* possible groups.

## 6. Prostate Cancer Classification

The first real-world data set we looked at is a prostate cancer set that consists of two tasks. The first data set from (Singh et al., 2002) is made up of laser intensity images from microarrays. The RMA normalization was used to produce gene expression values from these images. The second data set from (Welsh et al., 2001) is already in the form of gene expression values. Although the collection techniques for both data sets were different, they share 12600 genes which are used as features in this experiment. We used the same experimental setup as in (Zhang et al., 2010), i.e. we used 70% of each task as training set. Similar to the synthetic experiments we compared different approaches to solve the classification problem. The results of 20 cross-validation splits are shown in Figure 9 where we compared the prediction performance of the pooled data set, the $\ell_{1,\infty}$, $\ell_{1,3}$, $\ell_{1,2}$ and $\ell_{1,1.5}$ norm regularization and we compared all these multi-task learning methods with single-task learning $\ell_1$. Pooling the data yielded as bad results as in the synthetic experiments. As in the synthetic experiments, the $\ell_{1,1.5}$ norm performs best and all Group-Lasso methods perform significantly better than single-task learning, where we tested the statistical significance with the Kruskal-Wallis rank-sum test and the Dunn post test with Bonferroni correction. Even with only two tasks, we observe that single task learning is significantly worse than multi-task learning.

[1] http://www.public.asu.edu/ jye02/Software/SLEP/index.htm

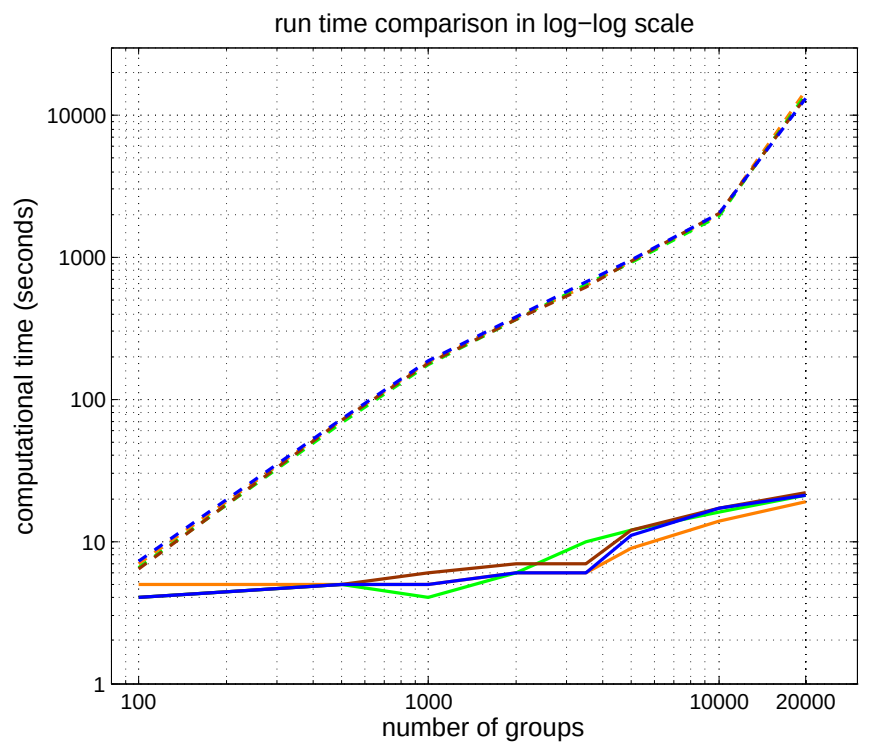

*Figure 8.* Run time in log-log-scale for our efficient active set algorithm (lines) and the algorithm proposed in (Liu & Ye, 2010) (dashed lines).

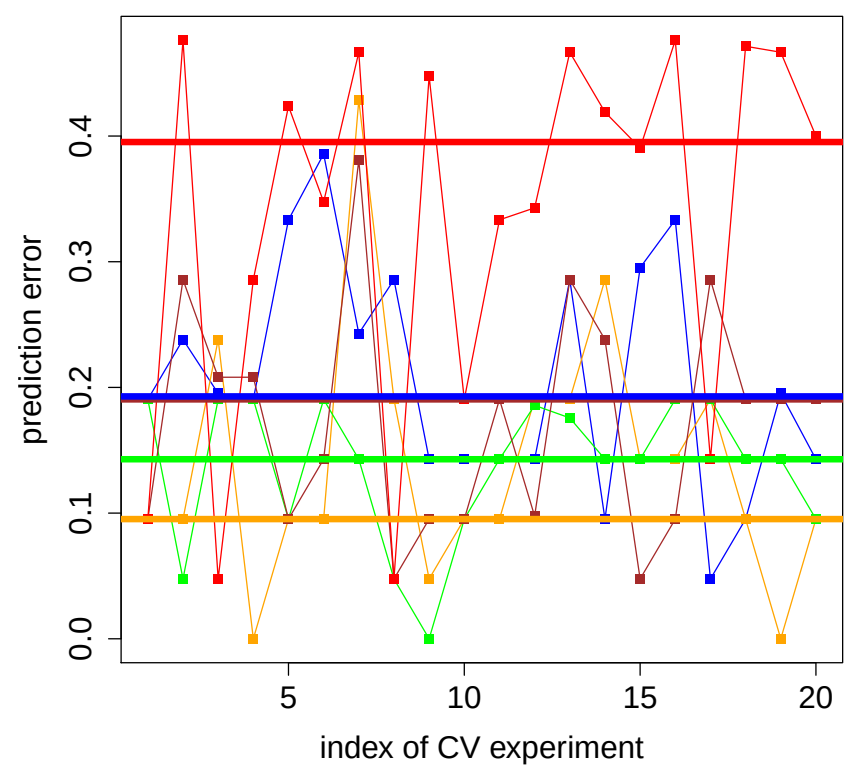

*Figure 9.* Classification error of the different Group-Lasso norms on the prostate cancer data set.

**MovieLens Data Set** In a second real world experiment, we applied different Group-Lasso methods on the MovieLens data set.[2] MovieLens contains 100,000 ratings for 1682 movies from 943 users. The genre information of the movies is used as features and the ratings of the users are in five-point scale (1, 2, 3, 4, 5). Every user defines a task, hence we have 943 tasks and 19 features, as we have the information about 19 movie genres. Similar as above, we compared different approaches to solve the learning problem. We conducted single-task learning and looked at different $\ell_{1,p}$ Group-Lasso methods. We see a statistical significant advantage of multi-task learning over single-task learning. Among the Group-Lasso methods, the very strong coupling of the $\ell_{1,\infty}$ norm yields the worst result. Between $\ell_{1,1.5}$ and the $\ell_{1,2}$ Group-Lasso there is no significant difference, but both show statistical significant advantages over all other methods. Figure 10 shows the results for the MovieLens data set, here we plotted single-task learning $\ell_1$, the $\ell_{1,2}$, $\ell_{1,\infty}$ and the $\ell_{1,1.5}$ Group-Lasso. One can see that single-task learning is significantly worse than multi-task learning and that the weak- and intermediate-coupling norms outperform the strong coupling norms.

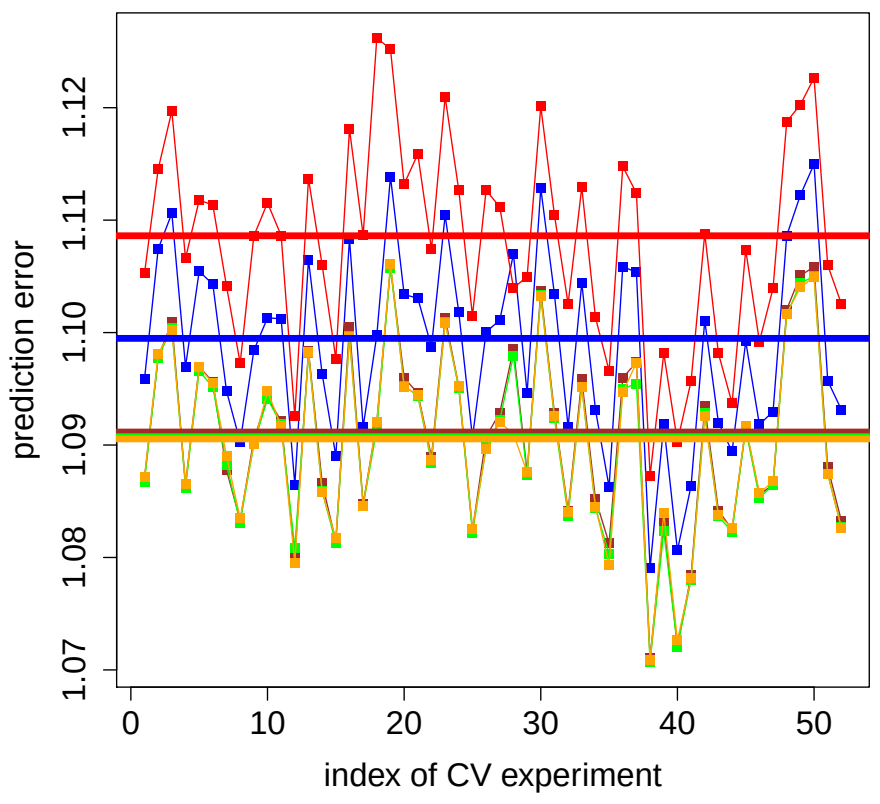

*Figure 10.* Prediction error of the different regularizers for the MovieLens data set: red curve: single $\ell_1$, orange curve: $\ell_{1,1.5}$, blue curve: $\ell_{1,\infty}$, green curve: $\ell_{1,2}$.

## 7. Conclusion

We have presented a unified characterization and a highly efficient active set algorithm for all $\ell_{1,p}$-variants of the Group-Lasso. With these results, we were able to compare Group-Lasso methods for different $p$-norms in large-scale experiments. To summarize, our contribution is threefold: (i) On the theoretical side, we characterized conditions for solutions for all $\ell_{1,p}$ Group-Lasso methods by way of subgradient calculus. (ii) We were able to present an active set algorithm that is applicable for all $\ell_{1,p}$ Group-Lasso methods. The main theoretical contribution consists in presenting a convergence proof of the interval bisection used to combine a constrained optimization problem and the Lagrangian form of an optimization problem in the inner optimization loop what leads to a fast update scheme. (iii) On the experimental side we compared the prediction performance of different Group-Lasso variants and demonstrated the computational efficiency of our method compared to an existing method. In a multi-task setting, where the different tasks are coupled via a Group-Lasso constraint, we examined the prediction performance of all $\ell_{1,p}$ variants. We compared the different methods on synthetic data as well as on two real-world data sets.

[2]The data is available at http://www.grouplens.org.

The prediction performance of the different Group-Lasso methods depends both on the coupling strength of the corresponding $\ell_p$ norms and on the systematic differences between the tasks. Our experiments indicate that both the very tight coupling of the high-$p$ norms with $p \gg 2$ and the too loose coupling of the low-$p$ norms with $p \ll 2$ significantly degrade the prediction performance. The weak-coupling norms for $p \in [1.5, 2]$ seem to be the best compromise between coupling strength and robustness against systematic differences between the tasks.